# 深層畳み込みニューラルネットワークを用いた画像のカラー化
# Image Colorization Using a Deep Convolutional Neural Network

グエン・トゥン/立命館大学大学院 情報理工学研究科，森和貴，ターウォンマット・ラック/立命館大学 情報理工学部

Tung Nguyen[1]/Graduate School of Information Science and Engineering, Ritsumeikan University, Kazuki Mori[2], Ruck Thawonmas[3]/College of Information Science and Engineering, Ritsumeikan University

*[1]is0149sk @ed.ritsumei.ac.jp, *[2]is0191kh@ed.ritsumei.ac.jp, *[3]ruck@is.ritsumei.ac.jp

Abstract: In this paper, we present a novel approach that uses deep learning techniques for colorizing grayscale images. By utilizing a pre-trained convolutional neural network, which is originally designed for image classification, we are able to separate content and style of different images and recombine them into a single image. We then propose a method that can add colors to a grayscale image by combining its content with style of a color image having semantic similarity with the grayscale one. As an application, to our knowledge the first of its kind, we use the proposed method to colorize images of ukiyo-e—a genre of Japanese painting—and obtain interesting results, showing the potential of this method in the growing field of computer-assisted art.

## 1. Introduction

Image colorization is the process of assigning colors to a grayscale image to make it more aesthetically appealing and perceptually meaningful. This is known to be a sophisticated task that often requires prior knowledge about the image content and manual adjustments in order to achieve artefact-free quality. Furthermore, since objects can be in different colors, there are many possible ways to assign colors to pixels in an image, which means there is no unique solution to this problem.

There are two main approaches for image colorization: one that requires user to assign colors to some regions and extends such information to the whole image, and another one that tries to learn the color of each pixel from a color image with similar content. In this paper, we use the latter approach; we extract the information about color from an image and transfer it to another image.

Recently, deep learning has gained increasing attention among researchers in the field of computer vision and image processing. As a typical technique, convolutional neural networks (CNNs) have been well-studied and successfully applied to several tasks such as image recognition, image reconstruction, image generation, etc. A CNN consists of multiple layers of small computational units that only process portions of the input image in a feed-forward fashion. Each layer is the result of applying various image filters, each of which extracts a certain feature of the input image, to the previous layer. Thus, each layer may contain useful information about the input image at different levels of abstraction.

With the evolution of computational resources, especially the computing power of GPUs, it has become possible to train very deep CNNs, and they have achieved some remarkable results recently. For example, a deep CNN (He et al., 2015) has surpassed human-level performance on ImageNet classification, or an adversarial network (Radford et al., 2015), in which two CNNs are trained simultaneously, is capable of generating plausible-looking images of many kinds of objects. These amazing successes of CNNs have motivated us to further investigate and explore their potential in the aforementioned image colorization problem.

## 2. Related work

Gatys et al. (2015) presented an application of deep neural networks that could learn to transfer style from an image to another one. Given an image the content of which is to be preserved (content image) and another image the style of which is to be transferred (style image), they passed both images into a pre-trained CNN and extracted the content representation and the style representation, respectively. They then did that to a noisy image and made changes to it until they obtained similar representations as of the content image and the style image. Indeed, this is an optimization problem in which the objective is to minimize the loss of reconstructing content and style simultaneously. As mentioned in their paper, they used gradient descent to solve this problem.

We follow Gatys et al.'s approach in our work, but use a different optimization method since gradient descent normally requires much fine-tuning of parameters to work well on a specific problem.

## 3. Methodology

In this paper, we combine content of a grayscale image and style of a color image, which results in colorizing the grayscale image. More details of our methodology are described below.

Considering a certain layer $l$ in the network, we denote the number of feature maps as $N_l$ and the size ($width \times height$) of a feature map in that layer as $M_l$. Furthermore, we denote the content image, the style image, and the noisy image as $p$, $a$, and $x$ respectively. The content loss is calculated by the following equation:

$$\mathcal{L}_{content}(p, x, l) = \frac{1}{2} \sum_{i,j} (P_{ij}^l - F_{ij}^l)^2,$$

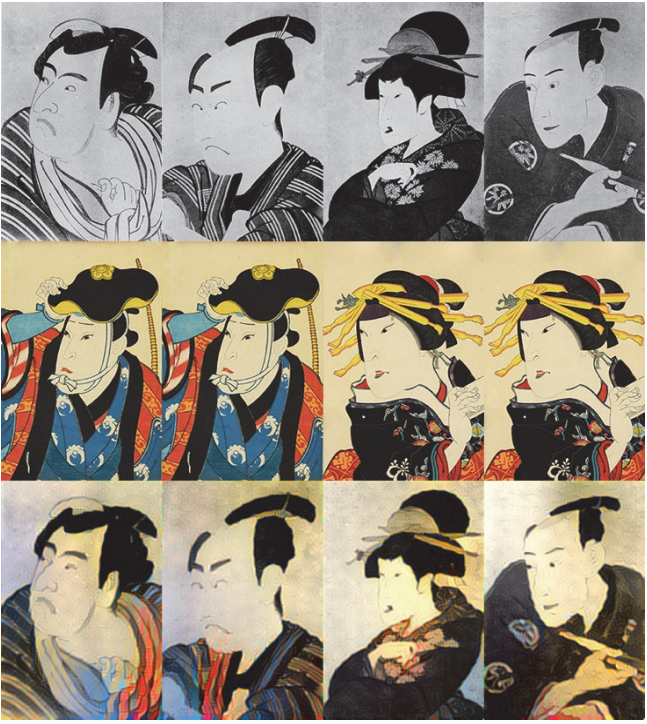

Figure 1: Examples of generated images. Top row: content images, middle row: style images, bottom row: colorized images.

where $P^l \in \mathbb{R}^{N_l \times M_l}$ and $F^l \in \mathbb{R}^{N_l \times M_l}$ are the content representations, i.e., the features of $p$ and $x$, respectively. On the other hand, the style representation of an image is given by the Gram matrix $G^l \in \mathbb{R}^{N_l \times N_l}$:

$$G^l = F^l(F^l)^T$$

The style loss at layer $l$ is calculated by

$$E_l = \frac{1}{4N_l^2 M_l^2} \sum_{i,j} \left(A_{ij}^l - G_{ij}^l\right)^2,$$

where $A^l$ and $G^l$ are the style representations of $a$ and $x$, respectively. However, as shown in Gatys et al.'s work, it is better to consider style losses at multiple layers, therefore the style loss function is defined as

$$\mathcal{L}_{style}(a, x) = \sum_{l=0}^{L} w_l E_l,$$

where $w_l$ is the corresponding weighting factor of $E_l$. Finally, the total loss function is defined as the weighted average of the content loss and the style loss:

$$\mathcal{L}_{total}(p, a, x) = \alpha \mathcal{L}_{content}(p, x) + \beta \mathcal{L}_{style}(a, x)$$

In order to improve optimization efficiency, we use L-BFGS—a quasi-Newton method that approximates the Broyden–Fletcher–Goldfarb–Shanno (BFGS) algorithm using a limited amount of computer memory—as an optimization method to find the minimum of $\mathcal{L}_{total}$. Moreover, we decrease the ratio $\beta/\alpha$ by 0.25% after each iteration, which is different from the related work, so that the colorization process does not give an implausible result by assigning too many colors to the original image.

As done in the related work, we also utilize the VGG-19 network (Simonyan and Zisserman, 2015) which was originally trained on the ImageNet dataset for image recognition. We use only layer 'conv4_2' for content reconstruction and 5 layers 'conv1_1', 'conv2_1', 'conv3_1', 'conv4_1', 'conv5_1' for style reconstruction. The weighting factor $w_l$ is set to $1/5$ for all layers in the style loss function.

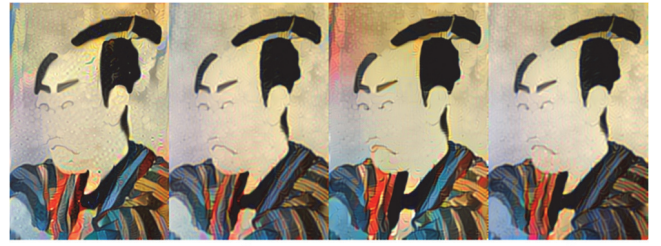

Figure 2: Images generated by different optimization methods. (a) SGD, fixed style weight. (b) SGD, decreasing style weight. (c) L-BFGS, fixed style weight. (d) L-BFGS, decreasing style weight (this work).

We applied our proposed method to colorize images of ukiyo-e, which is a genre of Japanese painting and represents one of the highpoints of Japanese cultural achievement. Figure 1 shows some examples of generated images, each of which was obtained after running 1000 iterations on a Torch implementation (Johnson, 2015). We also compared different optimization methods and illustrated the results in Fig. 2. For stochastic gradient descent (SGD), we had to do fine-tuning of parameters manually and showed only results generated with the best parameters we could find. In contrast, L-BFGS not only required no fine-tuning, but produced images that look slightly better. For both optimization methods, it can be seen that decreasing the style weight over iterations helped improve the visual quality of generated images.

## 4. Conclusions

In this paper, we presented a reliable method for colorizing grayscale images that uses a CNN to extract color information from an image and transfer to another image. We showed examples of plausible-looking generated images. Our results indicate that the presented method can be used as a creativity tool to assist human artists in near future.

## References

K. Simonyan and A. Zisserman, 2015. Very Deep Convolutional Networks for Large-Scale Image Recognition. *International Conference on Learning Representations 2015* (*arXiv:1409.1556v6*).

K. He, X. Zhang, S. Ren, and J. Sun, 2015. Delving Deep into Rectifiers: Surpassing Human-Level Performance on ImageNet Classification. *IEEE International Conference on Computer Vision 2015*, pp. 1026-1034.

L. A. Gatys, A. S. Ecker, and M. Bethge, 2015. A Neural Algorithm of Artistic Style. *arXiv:1508.06576v2*.

A. Radford, L. Metz, and S. Chintala, 2016. Unsupervised Representation Learning with Deep Convolutional Generative Adversarial Networks. Under review for *International Conference on Learning Representations 2016* (*arXiv:1511.06434v2*).

J. Johnson, 2015. Torch implementation of neural style algorithm. https://github.com/jcjohnson/neural-style (last accessed on February 3, 2016).